\pdfoutput=1

\documentclass[11pt]{article}

\usepackage{acl}

\usepackage{times}
\usepackage{latexsym}

\usepackage[T1]{fontenc}
\usepackage{amsmath}
\usepackage{amssymb}

\usepackage[utf8]{inputenc}

\usepackage{microtype}
\usepackage{booktabs}
\usepackage{todonotes}
\usepackage{xspace}
\usepackage{cleveref}
\usepackage{multirow}
\usepackage{subcaption}

\usepackage{todonotes}

\newcommand\blankfootnote[1]{%
  \let\thefootnote\relax\footnotetext{#1}%
  \let\thefootnote\svthefootnote%
}

\usepackage{color, colortbl}
\usepackage{xcolor, soul}
	
\definecolor{Gray}{gray}{0.1}
\title{Adapting Pretrained Text-to-Text Models for Long Text Sequences}

\author{Wenhan Xiong$^\ast$, Anchit Gupta$^\ast$, Shubham Toshniwal,  \\ {\bf Yashar Mehdad}, {\bf Wen-tau Yih} \\ \\
 Meta AI 
 \\
 {\small{\tt \{xwhan,anchit,shtoshni,mehdad,scottyih\}@fb.com}}
 }
\begin{document}
\maketitle
\begin{abstract}


We present an empirical study of adapting an existing pretrained text-to-text model for long-sequence inputs. Through a comprehensive study along three axes of the pretraining pipeline -- model architecture, optimization objective, and pretraining corpus, we propose an effective recipe to build long-context models from existing short-context models. Specifically, we replace the full attention in transformers with \textit{pooling-augmented blockwise attention}, and pretrain the model with a masked-span prediction task with spans of varying lengths. In terms of the pretraining corpus, we find that using randomly concatenated short-documents from a large open-domain corpus results in better performance than using existing long document corpora, which are typically limited in their domain coverage. With these findings, we build a long-context model that achieves competitive performance on long-text QA tasks and establishes the new state of the art on \emph{five} long-text summarization datasets, often outperforming previous methods with larger model sizes.\blankfootnote{$^\star$~Equal Contribution.}\blankfootnote{$^\_$~Our code has been released at \url{https://github.com/facebookresearch/bart_ls}.}

\end{abstract}

\section{Introduction}

NLP applications like summarization and question answering often require processing long text sequences. While there have been tremendous empirical breakthroughs~\citep{Transformers, devlin-etal-2019-bert} from large pretrained language models (PLMs), most of these successes have been confined to \emph{short-context} tasks~\citep{rajpurkar-etal-2016-squad,glue}. On long-context NLP benchmarks~\citep{narrativeqa, qmsum, quality}, where the input sequences are often longer than 10,000 tokens, there is still a significant gap between human performance and the state-of-the-art models.

Extending the success of PLMs to long texts is nontrivial for the following reasons. First, the quadratic complexity of self-attention makes it prohibitive to directly apply full-attention to long sequences. Any long-range architecture needs to be computationally efficient and at the same time capture long-distance dependency.\footnote{While there exists a long list of efficient attention variants~\citep{attention_survey}, their efficacy is only validated in synthetic or small-scale experiments and it is unknown whether these variants are scalable and suitable for large-scale pretraining for natural language~\citep{SimpleLA, scaling_laws}.} 
Second, the training objectives used by existing PLMs
have largely focused on short text and have not been well-studied for long-context scenarios. For instance, BART~\citep{lewis-etal-2020-bart} pretraining involves reconstructing the whole corrupted input sequence, which is impractical for long sequences given the computational overhead of decoder-side attention. 
Additionally, while abundant short documents can be easily collected from web dumps 
to pretrain short-context models that work well across different domains, long documents are much scarcer and are often collected from specific domains as books or movie scripts~\citep{thepile}.
It is unknown whether the existing corpora are more effective for pretraining a versatile long-context model compared to using artificially constructed long texts. 

In this work, we conduct a thorough experimental study to find a
recipe for building
high-performing
long-context models. In contrast to a recent work~\citep{longt5} that pretrains a long-context model from scratch, we choose to adapt an existing short-text model for long texts with further pretraining. 
Our empirical results demonstrate the effectiveness of this strategy by achieving stronger performance on various downstream tasks, while saving on the high cost of pretraining from scratch.
More specifically, we explore three axes of the pretraining pipeline, namely \textit{efficient long-range model architectures}, \textit{long text corpora creation} and \textit{the choice of pretraining objectives}. Our main findings are summarized as follows:

1) Among long-range mechanisms, such as global tokens and sliding-window attention, we find a simple pooling-augmented blockwise attention to be the most effective choice for various tasks. 

2) For the pretraining corpus, we surprisingly find that using randomly concatenated documents from a large open-domain corpus (CommonCrawl) performs better than using existing long-document corpora such as book collections. 

3) We experiment with various pretraining objectives including standard masked-span prediction~\citep{t5}, primary sentence prediction~\citep{pegasus}, and a novel model-based span prediction objective. While we find all of these objectives can bring gains
over models that are not pretrained on long texts, we consider the masked-span prediction objective (using both short and long spans) remains as the best choice, thanks to its simplicity and balanced effectiveness on both short-  and long-output tasks.

Using these findings, we build a strong
long-context text-to-text model 
that establishes new state-of-the-art on five long-text summarization tasks (with $>10\%$ relative ROUGE-2 improvements on three of the datasets) and achieves competitive performance on long-text QA tasks  despite its comparatively modest size. In the following two sections, we first describe the details of various choices considered along the three axes and then present the corresponding results and analysis.

\section{Model and Data}

\label{sec:background}

\subsection{Efficient Models for Long Sequences}
\label{sec:arch_desc}

Our model is based on a standard transformer with block-sparse self-attentions~\citep{Bigbird} on the encoder side. While various new architectures~\citep{linformer,performer,sru+,s4} have been proposed, we stick to the simple architecture for the following reasons: 1) it makes it easy to reuse existing pretraining pipelines, which are often highly optimized specifically for vanilla transformers, e.g., learning rate schedules, normalization layers, optimizers; 
2) using local attentions, where each token attends to only tokens in the local context, allows our model to reuse all the model parameters from existing PLMs, while other attention variants use different parameterizations that prohibit inheriting the weights of an existing pretrained model.

In addition to block attention, we investigate three mechanisms that enable long-range connections in the encoder: 

1) \textit{\textbf{Global-token mechanism}}: Previous work \citep{longt5,Bigbird,longformer} has proposed augmenting block-sparse attention with a small set of ``global tokens'' that attend to the entire sequence and hence enable long-range interactions in the encoder. Specifically, we mark the first 64 tokens in each attention block as global tokens and share the projection matrices for both the global and regular tokens. This mechanism has proven effective in encoder-only models, especially for question answering tasks as shown by the aforementioned methods.

2) \textit{\textbf{Overlapping (strided) attention windows}}: Sliding-attention with overlap is a straightforward way to introduce long-range connections in local attention models. As we stack the layers in the encoder, the receptive field of each token would increase exponentially. For example, \cite{longformer} use the stride of one token and each token attends to an equal number of tokens from both sides. We develop a simpler and faster block-wise version which makes the parallelization easier to implement; namely, tokens in each block will attend to all the tokens inside the block, and half of the tokens from its immediate left and right blocks.

\begin{figure}[t]
\centering
\includegraphics[width=0.5\linewidth]{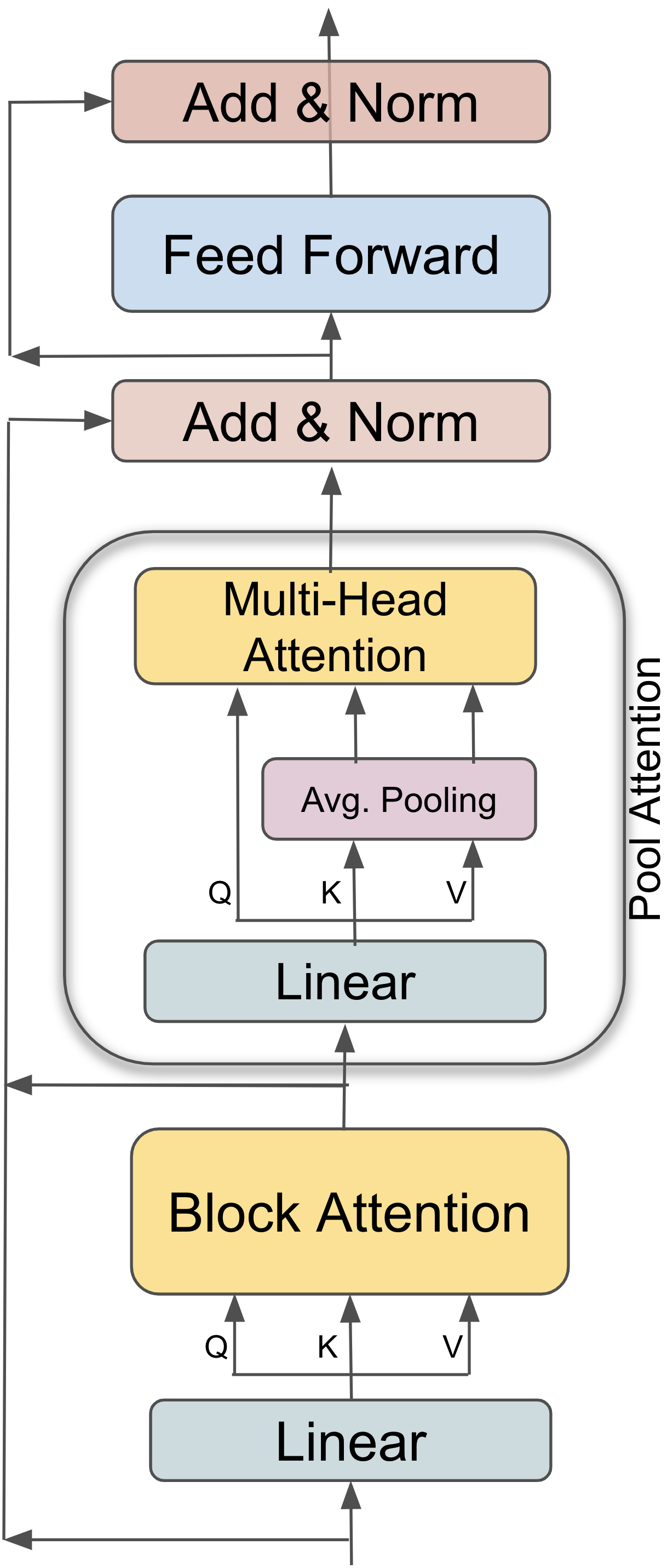}
\caption{The pooling augmented self-attention layer. The pooling attention parameters marked separately are newly introduced and randomly initialized.} 
\vspace{-10pt}
\label{pooling_layer}
\end{figure}

3) \textit{\textbf{Pooling layers}}: Recent work~\citep{poolingformer, topdown} has explored using pooling operations to reduce the number of \textit{key} and \textit{value} states in transformers. We implement a simpler version that only requires standard average pooling operations.
All illustration of the pooling-augmented attention layer is shown in Figure~\ref{pooling_layer}. Specifically, in the top n layers of the transformer encoder, we add a second attention module which takes as input the hidden states output by the $i$th block self-attention layer $\mathbf{X_i} \in \mathbb{R}^{L \times h}$, where $L$ is the sequence length and $h$ is the size of the hidden states. As in the vanilla attention layers, $\mathbf{X_i}$ is first projected to create the \textit{key, query, value} matrices $\mathbf{Q}_i^p, \mathbf{K}_i^p, \mathbf{V}_i^p \in \mathbb{R}^{L \times h}$.\footnote{The projection layers to create these matrices are not used in existing pretrained models and will be randomly initialized before further pretraining}
We first average pool the $\mathbf{K}_i^p$ and $\mathbf{V}_i^p$ sequences, with a fixed kernel/stride size, into smaller lengths $\tilde{\mathbf{V}}_i^p$, $\tilde{\mathbf{K}}_i^p$ $\in \mathbb{R}^{\tilde{L} \times h}$, where $\tilde{L} \ll L$. We then apply standard attention using $\mathbf{Q}_i^p$, $\tilde{\mathbf{K}}_i^p$ and $\tilde{\mathbf{V}}_i^p$ resulting in $O(L \times \tilde{L})$ complexity. The output of the pooling layers is added with $\mathbf{X_i}$ to form a residual connection. 

We compare these variants via the performance on downstream long-sequence tasks in Sec~\ref{sec:arch_analysis}. 

\subsection{Pretraining Corpus}
\label{subsec:pretrain_corpus}

The choice of the corpus has a significant impact on the downstream results. We consider long documents from formal text domains, including Books3~\citep{thepile}, STORIES~\citep{stories}, RealNews~\cite{Realnews}; and long dialogues including MediaSum~\citep{mediasum} and OpenSubtitles~\citep{opensub}. While collecting a long-document corpus seems to be a natural choice for long-sequence downstream tasks, as they are more likely to include long-range dependencies than common short texts on the internet, pretraining only on these datasets also brings the risk of overfitting to specific domains, instead of achieving consistent gains on a range of tasks. 
Thus, we also consider a general-domain corpus -- C4 as used by T5~\citep{t5}.  
Additionally, instead of using randomly concatenated sequences, we also tried to concatenate semantically similar C4 documents (using similarity metric learned by dense retrieval models) with the hope that the model can learn to capture more long-range dependencies across relevant documents. 
We discuss the effects of these corpus variants in Sec~\ref{sec:corpus}.

\subsection{Pretraining Objectives}
A variety of self-supervised pre-training objectives have been proposed for sequence-to-sequence models~\citep{lewis-etal-2020-bart, t5, longt5}.
In the long document setting, we ideally seek an objective that promotes long-range reasoning ability in the model. We investigate the following different pretraining objectives and the effect of input length during pretraining.

1) \textit{\textbf{T5 Span Denoising}}: 
Applying BART's denoising objective to long sequences is computationally expensive as it requires reconstructing the entire input and incurs significant computation overhead on the decoder-side attention.
Moreover, reconstructing the entire input would be at odds with most downstream tasks such as question-answering and summarization, which require generating shorter text.
Thus, we adopt T5-style denoising for pretraining our model, i.e., we randomly pick a set of spans in the input sequence as the decoding target and mark them with special sentinel tokens. The model is then trained to generate the uncorrupted spans. This objective is readily applicable to long documents as we can control both the length and the number of spans. We experiment with both fixed span lengths as in \citep{t5}, and also mixed span lengths with both short and long spans, with which we hope the model is able to perform well on a range of tasks requiring differing output lengths.

2) \textit{\textbf{Pegasus -- Primary Sentence Prediction}}: Originally proposed for summarization pretraining in \citep{pegasus} and recently used for long documents by \citet{longt5}, this objective identifies and masks out a set of principle sentences, i.e., sentences with a high ROUGE
score with the rest of the document. The model is then trained to generate these principle sentences. The output length can be controlled by choosing the number of principle sentences to mask.

3) \textit{\textbf{Model-based Denoising}}: Apart from randomly selecting the decoding targets, we also explore a novel model-based objective. 
Here we use a separate encoder-only model (with local attention) to select decoding targets for the sequence-to-sequence model. This approach is inspired by ELECTRA~\citep{electra} and we hope the prediction loss of the encoder-only model can be a good proxy to select spans that require long-range dependencies to predict. Specifically, we first mask a larger number of tokens (5,120 tokens instead of 1,024) in the input sequence. We then apply an encoder-only masked language model to recover the masked spans. Based on the losses of the masked language model, we only keep the top 20\% hard spans to train the text-to-text model.
The encoder-only model can either be frozen or jointly trained with the sequence-to-sequence model.

\section{Experiments}
\label{sec:exp}

\subsection{Downstream Tasks \& Finetuning Setup}
\label{sec:setup}

We evaluate the models on six summarization datasets and four QA datasets. The summarization datasets are from formal domains, including \textbf{GovReport}~\citep{govreport}, \textbf{ArXiv} \& \textbf{PubMed}~\citep{arxiv_pubmed} and \textbf{BookSum Chapters}~\citep{booksum}; or informal conversational domains
, such as TVMegaSite \& ForeverDreaming~\citep{summscreen}. For \textbf{QA}, we consider \textbf{Qasper}~\citep{qasper}, which contains questions over NLP papers; \textbf{QMSum}\footnote{QMSum is proposed as a "query-based summarization" dataset. We consider it as a special case of QA as our model uses the same input format for QMSum and other QA datasets.}~\citep{qmsum}, longform QA over meeting scripts, and two QA datasets on books: \textbf{QuALITY}~\citep{quality} and \textbf{NarrativeQA}~\citep{narrativeqa}. 

We finetune the pretrained model with a maximum of 16,384 tokens. For long-sequence QA tasks, we adopt the input format as used by the state-of-the-art open-domain QA system~\citep{fid} and the query-based summarization model~\citep{qmsum_trick}. Specifically, we repeat the question/query at the start of each attention block. 
For finetuning, we utilize the robust finetuning technique proposed by~\citet{r3f}. We also conduct a grid search over finetuning  hyperparameters, such as learning rate and dropout rate, the details of which are presented in Table~\ref{tab:finetune_hyper} in Appendix~\ref{sec:appendix}. We report ROUGE\footnote{\url{https://github.com/pltrdy/files2rouge}} scores for summarization datasets, and for QA we use Exact Match (EM) scores for datasets with short answers and F1 scores for datasets with long answers.

\subsection{Effect of Architectures}
\label{sec:arch_analysis}

\begin{table*}
\small
\centering
    \begin{tabular}{l|cc|cc|cc|c|c}
    \toprule
    \multirow{2}{*}{Models} &  \multicolumn{2}{c|}{GovReport} & \multicolumn{2}{c|}{ArXiv} & \multicolumn{2}{c|}{QMSum} & Qasper & QuALITY \\ 
    & R-1 & R-L & R-1 & R-L & R-1 & R-L & Ans F1 & Ans EM \\
    \midrule
    block-attn baseline.    & 60.5 & 57.5 & 49.0 & 44.2 & 35.2 & 30.4 & 28.0 & 31.6 \\
    \midrule
    + attn window overlaps & 60.6 & 57.6 & 49.0 & 44.3 & 34.8 & 30.2 &  28.0 & 31.6 \\
    + global tokens        & 60.3 & 57.3 & 49.1 & 44.3 & 35.4 & 30.7 & 29.8 & 32.5 \\
    + pooling layers       & \textbf{61.0} & \textbf{58.1} & \textbf{49.1} & 44.3 & 35.9 & 31.2 & \textbf{30.6} & \textbf{32.9} \\
     \bottomrule
    \end{tabular}

\caption{Ablation of different long-range mechanisms using \textit{base-size} models.}
\label{tab:arch_ablation}
\end{table*}

We study the effectiveness of different model choices before launching the resource-consuming pretraining. We first initialize a base-size block-attention model using BART's weights. We augment the model with three additional long-range mechanisms, as described in Sec~\ref{sec:arch_desc}. Note that only the pooling layers introduce additional parameters that will be randomly initialized. Table~\ref{tab:arch_ablation} shows the results on both QA and summarization tasks. For the \textit{global-token} mechanism, we mark the first $64$ tokens of each block as global tokens. We see that pooling layers produce the most consistent improvements even for GovReport, where the baseline already achieves strong numbers. Consistent with a prior study on encoder-only models~\citep{SimpleLA}, attention window overlaps fail to produce further improvements over the disjoint block-attention layers. Adding global tokens consistently helps on QA tasks but not on summarization tasks. We hypothesize that in encoder-decoder models, the cross-attention can offset the effect of global tokens, as each decoding position has access to all input tokens' representations. When finetuning our final pretrained model, we also try to combine global tokens with pooling layers for QA tasks, but we do not observe further improvements.

\subsection{Effect of Pretraining Corpus}
\label{sec:corpus}

\begin{table*}
\small
\centering
    \begin{tabular}{l|c|c|c|c}
    \toprule
    
    \multirow{2}{*}{Models} & QMSum & Qasper & QuALITY & NarrativeQA  \\ 
    & R-1 & Ans F1 & Ans EM & Ans F1  \\
    \midrule
    non-pretrain & 35.9 & 30.6 & \textbf{32.9} & 20.4\\
    \midrule
    Long corpus  & 34.7  & 29.9 & 31.3 & 21.2\\
    C4           & \textbf{36.3}  & \textbf{32.8} & 32.8 & \textbf{21.6} \\
    C4-linked   & 35.7  & 32.1 & 32.8 & 21.3 \\
     \bottomrule
    \end{tabular}
    
\caption{Effects of pretraining corpus. Base size models pretrained for 20k steps to avoid repetitions. \textbf{Long corpus}: \textit{Books3 + RealNews + STORIES + MediaSum + OpenSubtitles}; \textbf{C4}: \textit{randomly concatenated documents to form long sequences.}; \textbf{C4-linked}: \textit{concatenate related short documents using a retriever model.}}
\label{tab:corpus}
\end{table*}

\begin{figure}[t]
\centering
\includegraphics[width=\linewidth]{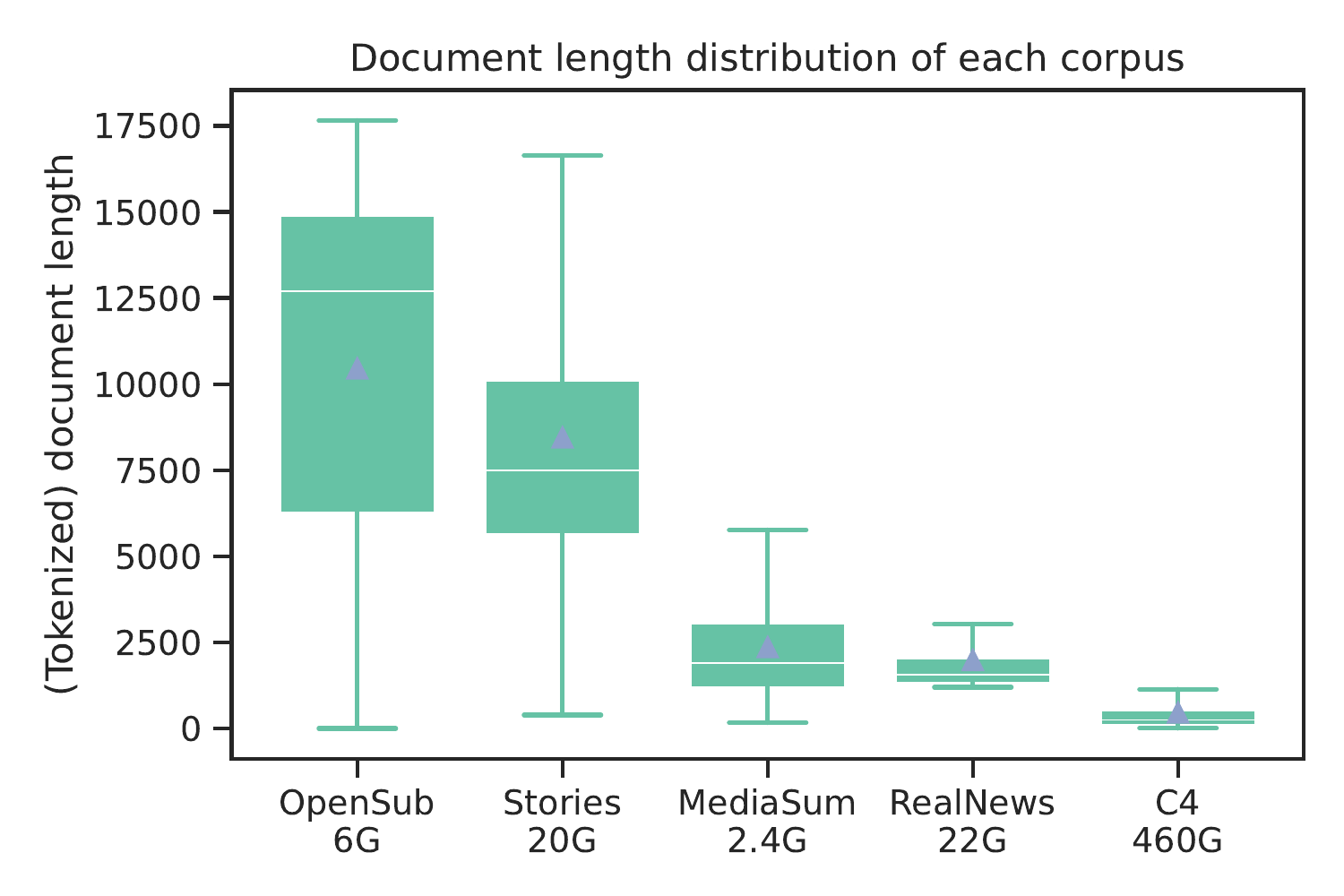}
\caption{Document length distribution of each source corpus. The sizes of each corpus (file sizes of tokenized texts) are also shown in the x-axis. The median and mean lengths are denoted via the while line and the triangle. We did not show the statistics of the Books3 corpus (60G) here as it has much longer documents with mean/medium over 100k tokens.}
\label{corpus_lens}
\vspace{-10pt}
\end{figure}

With the assumption that models should be exposed to as many long dependencies as possible at pretraining time, we initially tried to only pretrain the model with natural long documents that are collected from sources like books, news, and TV dialogues. However, we did not achieve consistent improvements with this corpus alone. Instead, we found it is important to include sufficient documents from diverse domains, even if those documents are mostly short sequences. We present our ablation analysis in Table~\ref{tab:corpus}. Here we reported results on small summarization datasets where the gaps are more visible. Note that the sizes of long-document corpora are usually smaller than open-domain corpus. To remove the size factor that affects model performance, we limit the pretraining steps such that the model does not see repeated examples from each corpus. In Figure~\ref{corpus_lens}, we show the length statistics of each document source. 

From the table, we see that pretraining on corpora that only have long documents, which are often from specific domains, hurts the downstream performance for most of the datasets, except for NarrativeQA, which is from a very close domain. On the other hand, pretraining on randomly concatenated C4 documents brings visible gains for most of the tasks. In addition to directly using concatenations of random C4 documents, we tried to assemble long sequences using semantically similar C4 documents, with the hope of creating more long-range connections in the pretraining sequences. For each document, we use a dense retrieval model~\citep{contriver} to find similar documents and concatenate them as long pretraining sequences. We denote this corpus as ``\textit{C4-linked}". However, this new corpus is either similar or worse compared to directly using C4. We conjecture that it is because the retrieved documents may contain redundant information, making some of the masked spans trivial to predict --- the training perplexity after 100k updates on ``\textit{C4-linked}" is significantly lower than that on the original C4 corpus (10.5~vs~12.2). Example sequences and details on how to build this corpus can be found in the Appendix.

\subsection{Effect of Pretraining Objectives}

\begin{table}
\small
\centering
    \begin{tabular}{l|cc}
    \toprule
    
    \multirow{2}{*}{Models} & QMSum & Qasper  \\ 
     & R-1 & Ans F1 \\
    \midrule
     
    no-pretraining  & 35.9  & 30.6 \\
    \midrule
    + T5 avg span\_len 5 - 8k & 36.7  & 32.9  \\
    + T5 avg span\_len 5 - 16k & 37.0 & 34.6 \\
    + T5 mixed span\_len & 37.0 & \textbf{35.4} \\
    + pegasus   & \textbf{37.4} & 34.4 \\
    + model-based & 37.0 & 32.5 \\
     \bottomrule
    \end{tabular}
    
\caption{Ablation of different pre-training objectives on C4 corpus}
\vspace{-10pt}
\label{tab:obj}
\end{table}

We compare the effects of different pretraining objectives in Table~\ref{tab:obj}. The generation targets are usually paragraph-length for QMSum, while Qasper expects the model to predict spans or single sentences most of the time. All the models are pretrained for $100k$ updates on the C4 corpus. To investigate the effect of pretraining sequence length, we compare the $16k$ model with a model pretrained with $8k$ sequence length. We double the batch size for the $8k$ length pretraining such that the input tokens in each batch stays the same. We also increase the masking ratio for the $8k$ model to $1/8$ so that the decoding sequence length remains 1,024. Note that under this setting, pretraining with $8k$-length batches is a bit slower compared to the $16k$ batches due to the decoder-side self-attention.

\paragraph{Pretraining with longer sequences is useful.} While a prior work~\citep{longt5} pretrains their model with sequences shorter than downstream tasks, we find it is generally better to directly pretrain with longer sequences. In terms of convergence rate, we find pretraining with 8k and 16k sequences are similar (the loss curves can be found in Appendix~\ref{sec:appendix}). For downstream results, we find that training with longer sequences lengths is indeed helpful for low-resource datasets --- QMSum and Qasper are both small with a few thousand examples (\textit{T5 avg span\_len 5 - 8k} vs \textit{T5 avg span\_len 5 - 16k}). We also find using a range of short spans (\textit{mixed span\_len}) tends to give more gains on QA tasks.

\paragraph{Alternative objectives works similar as random masking.} While the \textbf{Pegasus} objective is effective for summarization, we do not find it to be consistently better than T5 denoising. It also incurs more data processing costs compared to T5's random masking. We also find that model-based denoising fails to yield better performance than random denoising, even though it introduces a harder pretraining task, i.e., larger training losses. We conjecture that, while this objective might provide more training signals that are related to long-range dependencies, it can also introduce noisy supervision, which is harmful for the model to learn a wide range of language understanding skills. 

\subsection{Main Results}

\begin{table*}[ht!]
\small
\centering
    \begin{tabular}{l|c|ccc|ccc|ccc|ccc}
    \toprule
    \multirow{2}{*}{Model} & \multirow{2}{*}{\# Param}  &  \multicolumn{3}{c|}{GovReport} & \multicolumn{3}{c}{BookSum} & \multicolumn{3}{c|}{ArXiv} & \multicolumn{3}{c}{PubMed}  \\ 
    & & R-1 & R-2 & R-L & R-1 & R-2 & R-L & R-1 & R-2 & R-L & R-1 & R-2 & R-L \\
    \midrule
    
    
    
    
    BigBird    & 580M & - & - & - & 31.8 & 6.5 & 14.2 & 46.6 & 19.0 & 41.8 & 46.3 & 20.7 & 42.3  \\
    LED & 460M & 59.4 & 26.5 & 56.6 & 32.8 & 7.5 & 14.6 & 46.6 & 19.6 & 41.8 & 47.0 & 20.2 & 42.9 \\
    PageSum     & 440M & 59.9 & 27.2 & 57.1 & - & - & - & 49.7 & 21.1 & 44.7 & 48.2 & 21.1 & 44.3 \\
    BART-Hepos  & 440M & 56.9 & 22.6 & 53.8 & - & - & - & 48.2 & 20.3 & 41.8 & 48.1 & 21.1 & 42.7\\
    \rowcolor{gray!30}
    \texttt{DYLE}         & 525M & 61.0 & 28.8 & 57.8 & - & - & - &46.4 & 18.0 & 41.5 & - & - & - \\
    
    LongT5-large & 750M &  - & - & - & - & - & - & 48.3 & 21.6 & 44.1 &  50.0 & 24.7 & 46.5 \\
    
    LongT5-xl    & 3B  & - & - & - & - & - & - & 48.4 & 21.9 & 44.3 & 50.2 & \textbf{24.8} & \textbf{46.7} \\
    
    Top-down (AvgP) & 460M & - & - & - & 37.9 & 9.1 & 18.0 & 48.7 & 20.7 & 43.9 & 48.3 & 21.4 & 44.2  \\
    \rowcolor{gray!30}
    Top-down (AdaP) & 660M$^{*}$ & - & - & - & 38.3 & 9.2 & 18.1 & \textbf{51.0} & 21.9 & \textbf{45.6} & \textbf{51.1} & 23.3 & 46.5  \\
    \midrule

    BART-LS  & 440M & \textbf{62.0} & \textbf{30.9} & \textbf{59.2} & \textbf{38.5} & \textbf{10.3} & \textbf{36.4}$^{\dagger}$ & 50.2 & \textbf{22.1} & 45.4 & 50.3 & 24.3 & 46.3 \\
    \bottomrule
    \end{tabular}
\sethlcolor{lightgray}
\caption{Results on long-document summarization. Pipelined approaches are highlighted in \hl{gray}. LED's results on GovReport are taken from PageSum~\citep{pagesum}. $^{*}$: The AdaPool version of the Top-Down model requires an additional encoder model to predict the weights in its pooling layers. $^{\dagger}$: The baseline R-L scores on BookSum-Chapters are taken from~\citet{topdown} and may not be directly comparable due to different ROUGE scripts.}
\label{tab:formal_doc_summ}
\end{table*}

\paragraph{Best Model Configuration.} Following the analysis of base-size models, we pretrain a large-size model with the best configuration, which consists of (a) block attention and pooling layer augmentations applied to the vanilla Transformer architecture, (b) long-sequence training data batches formed by randomly concatenating the documents from C4 corpus, and (c) T5 denoising loss with a mix of short and long spans as the training loss. We pretrain the model for 100K steps. Our model is denoted as ``BART-LS" in the following sections.

\begin{table*}
\small
\centering
    \begin{tabular}{l|ccc|ccc}
    \toprule
    \multirow{2}{*}{Model} & \multicolumn{3}{c|}{TVMegaSite} & \multicolumn{3}{c}{ForeverDreaming} \\ 
    &  R-1 & R-2 & R-L & R-1 & R-2 & R-L \\
    \midrule
    BART-large & 43.5 & 10.3 & 41.4 & 33.8 & 7.5 & 29.1 \\
    DialogLM &  45.6 & 10.8 & 43.3 & 35.8 & 8.3 & 30.8 \\
     
    Top-down (AvgPool) & 49.3 & 14.4 & 47.5 & 35.8 & 8.9 & 31.1 \\
    Top-down (AdaPool) & 51.0 & 14.7 & 49.0 & 36.8 & 9.2 & 31.1 \\
    \midrule

    BART-LS w/o pretrain & 50.9 & 14.5 & 48.9 & 37.1 & 9.6 & 32.5 \\
    \midrule
    BART-LS  & \textbf{51.8} & \textbf{17.2} & \textbf{50.0} & \textbf{39.1} & \textbf{10.7} &
    \textbf{33.5} \\
    \bottomrule
    \end{tabular}
\caption{Results of on long dialogue (scripts from TV series) and narrative summarization.}
\label{tab:summscreen}
\end{table*}

\subsubsection{Summarization}

 Table~\ref{tab:formal_doc_summ} shows results of formal long-document summarization. We first compare our model with models that directly reuse existing models' weights without further pretraining and newly introduce parameters. Apart from BigBird and LED that simply use encoder-side local attention to allow existing PLM to take longer context, we also consider more recent baselines including PageSum~\citep{pagesum} which investigates the locality bias on both encoder and decoder side; BART\-Hepos~\citep{govreport} which applies head-wise cross-attentions; \texttt{DYLE}~\citep{dyle} which combines a context extractor with a generator that only takes short text as input, and uses a complex training pipeline to provide supervision for the extractor. Our model outperforms BigBird, LED, and BART\-Hepos by a large margin. With simple sequence-to-sequence finetuning, our model also consistently outperforms PageSum and \texttt{DYLE} which are specifically designed for summarization tasks. Note that PageSum proposes the idea of using a weighted combination of multiple decoder predictions (corresponding to taking the encodings of different parts of the input sequences as inputs), which could be orthogonal to our method. 

Compared to LongT5 (large and xl sizes), our model achieves stronger performance on ArXiv and stays on-par on PubMed, even with much fewer parameters. The recently proposed Top-Down Transformer~\citep{topdown} applies a similar pooling operation at the finetuning stage. Our model architecture is similar to their ``Average Pooling`` variant but simpler to implement. With the proposed pretraining method, our model outperforms ``Top-down (AvgP)" on all tasks. Apart from ``Top-down (AvgP)", the authors also proposes a more advanced pooling layer that uses the token importance predicted by another encoder-only model to aggregate the hidden states for each pooling operation, i.e., ``Top-down (AdaP)". While this method should be orthogonal to our model during finetuning, we find the model-based adaptive pooling hard to replicate. Our model matches the performance of ``Top-down (AdaP)" performance on ArXiv and PubMed in terms of R-2/R-L, and surpass their results on BookSum.

In contrast to formal documents, dialogue texts, especially multi-person conversations, can be noisier, more unstructured, and cover more diverse topics within each document. We test our model on two summarization datasets collected from popular TV series~\citep{summscreen}. As shown in Table~\ref{tab:summscreen}, our model achieves even stronger relative gains compared to formal-domain datasets. Note that DialogLM~\citep{dialogLM} is specifically designed for the dialog domain and further pretrained a PLM checkpoint on dialog corpus. The large improvements over their results again suggest the importance of pretraining with open-domain corpus.  

\begin{table}[t]
\small
\centering
    \begin{tabular}{l|c|c|c}
    \toprule
    \multirow{2}{*}{Model} & Qasper  & NarrativeQA & QuALITY\\ 
    & F1 & EM & EM-T/H \\
    \midrule
    LongT5-base & 46.6  & 23.0 & 37.9/36.6 \\
    LongT5-large & 53.3 & 27.2 & 40.6/38.6 \\
    LongT5-3B & 53.1 &  29.3 & 46.0/42.1 \\
    \midrule
    
    \rowcolor{gray!30}
    Block-BART (dev) & 38.1 & 24.1 & 35.7 \\
    \rowcolor{gray!30}
    BART-LS (dev) & 40.6 & 25.4 & 37.6\\
    \midrule
    BART-LS  & 48.7 & 26.2 & 37.8/34.0 \\
    \bottomrule
    \end{tabular}
\sethlcolor{lightgray}
\caption{Test results on QA tasks. LongT5's numbers are taken from the Scrolls benchmark~\citep{scrolls}. We also compare our model with a block-attention baseline that reuses BART's weights on the dev set, as shown in \hl{gray} rows. Not that our model's size is in between LongT5-base and LongT5 large.}
\vspace{-10pt}
\label{tab:qa}
\end{table}
\begin{table}
\small
\centering
    \begin{tabular}{l|ccc}
    \toprule
    \multirow{2}{*}{Model} & \multicolumn{3}{c}{QMSum} \\ 
    &  R-1 & R-2 & R-L \\
    \midrule
    BART-large & 32.2 & 8.0 & 27.7  \\
    DialogLM & 34.0 & 9.2 & 30.0 \\
    \texttt{DYLE} & 34.4 & 9.7 & 30.1 \\
    LED & 34.2 & 10.3 & 30.0 \\
    SecEnc & 37.1 & 13.0 & 32.6 \\
   \rowcolor{gray!30}
    SecEnc-W & 37.8 & 13.4 & 33.4 \\
    \midrule
    Block-BART (ours) & 36.6 & 12.1 & 32.4 \\
    \midrule
    BART-LS  & \textbf{37.9} & 12.1 & 33.1 \\
    \bottomrule
    \end{tabular}
    
\sethlcolor{lightgray}
\caption{Results on query-based meeting summarization (QMSum). The highlighted row indicates additional data has been used for training.}
\label{tab:qmsum}
\end{table}

\begin{figure*}[h]
\centering
    \begin{subfigure}[t]{0.45\textwidth}
        \centering
        \includegraphics[width=\textwidth]{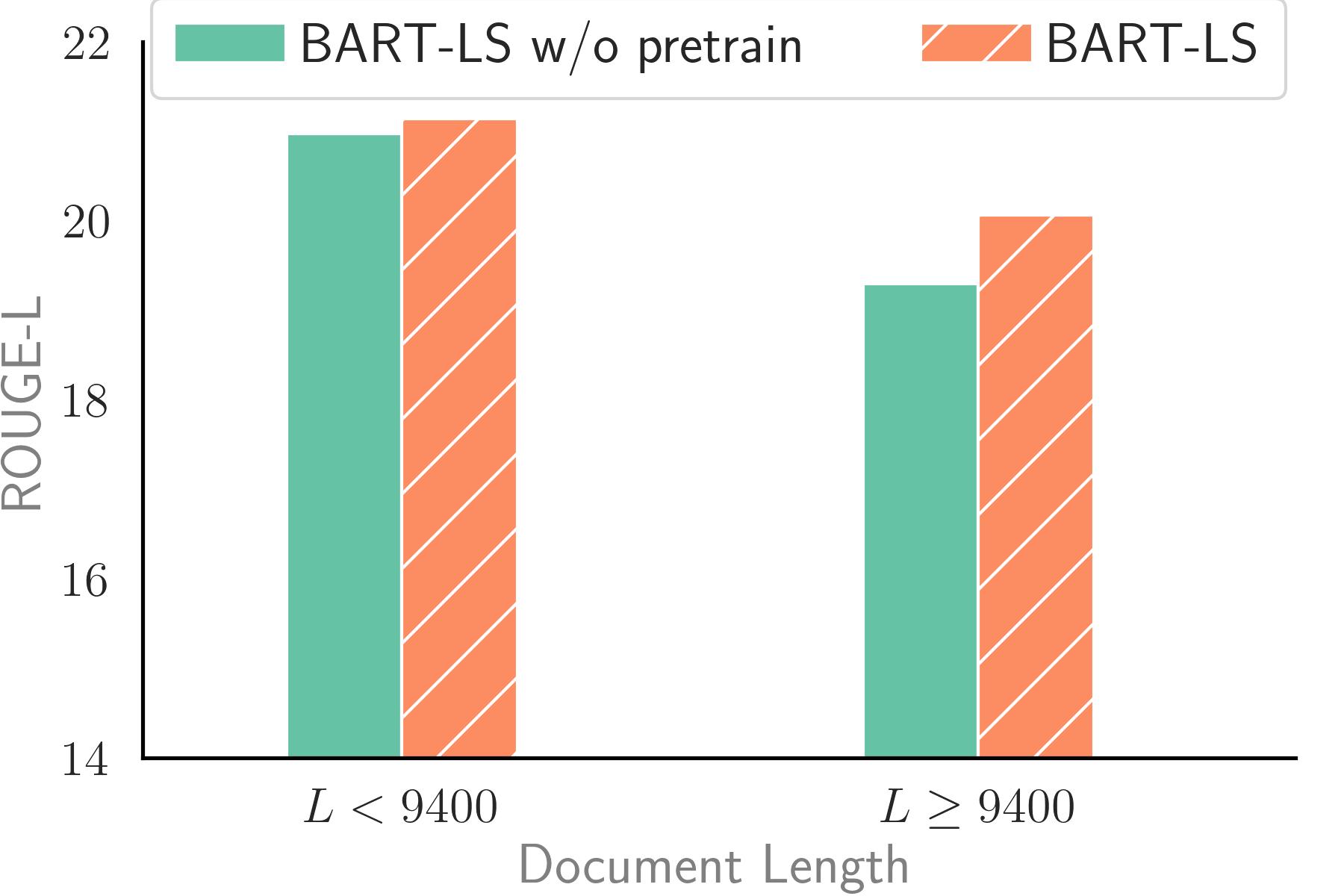}
        \caption{SummScreen}
    \end{subfigure}%
    ~ 
    \begin{subfigure}[t]{0.45\textwidth}
        \centering
        \includegraphics[width=\textwidth]{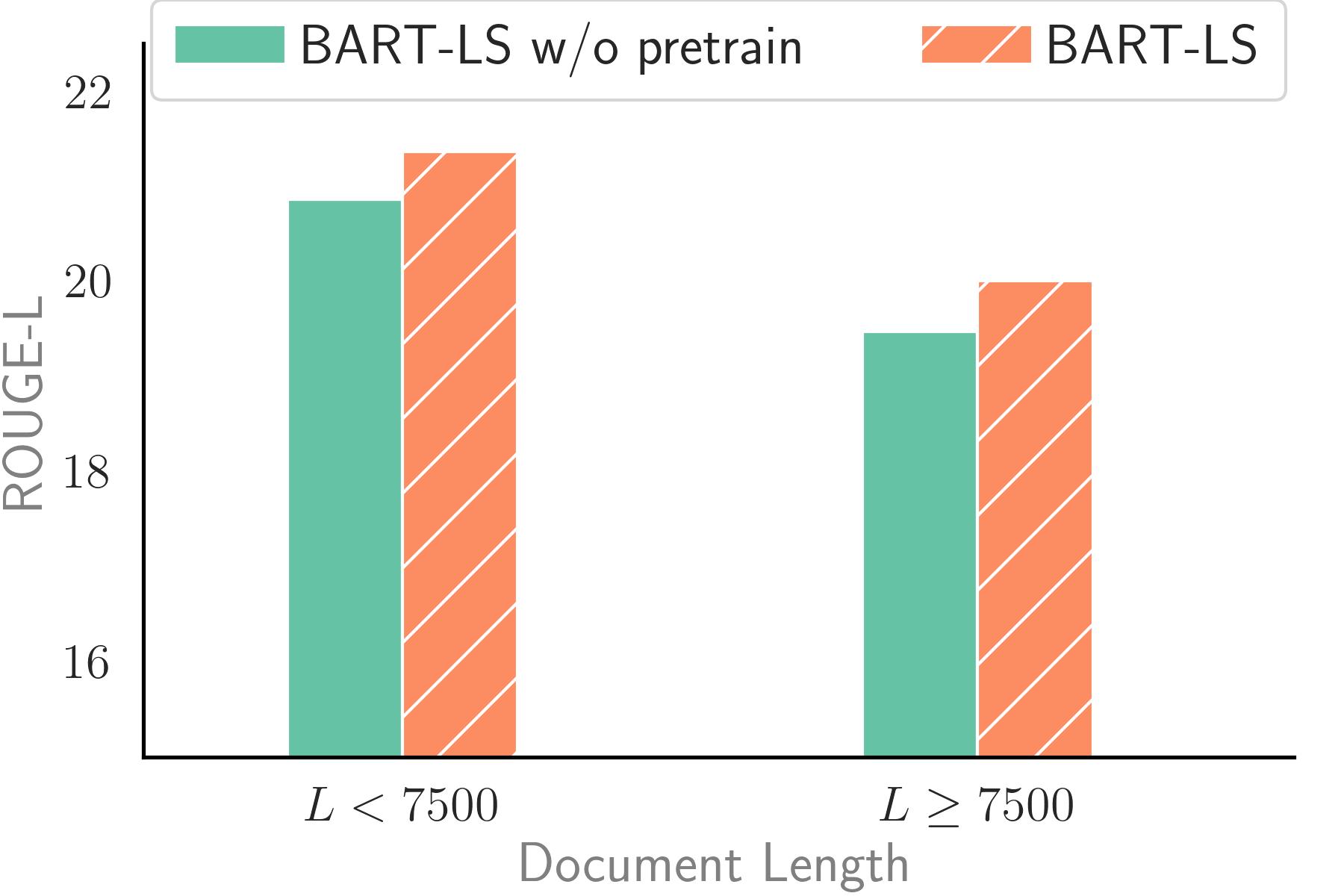}
        \caption{TVMegaSite}
    \end{subfigure}
    \caption{ROUGE-L scores as a function of source document length for the base model and the best model for two summarization datasets. }
    \label{fig:len_comp}
\end{figure*}
\subsubsection{QA and Query-Based Summarization}

As mentioned in Sec~\ref{sec:setup}, we use the same input format for finetuning QA tasks and query-based summarization. As there are no existing baselines of long models that reuse the weights of short-sequence models, we also report the performance of our implementation of block-attention BART. As shown in Table~\ref{tab:qmsum}, our model outperforms all previous methods that do not apply data augmentation. Here SecEnc~\citep{qmsum_trick} is also a block-attention version of BART -- it distribute overlapped texts (instead of disjoint text blocks) into each self-attention window and reuses the position embeddings of the first 1,024 tokens. On long-document QA datasets (as shown in Table~\ref{tab:qa}), our best model is consistently better than our block-attention baseline and is aligned with LongT5 scaling laws -- our model's size is between the base and large versions of LongT5. We believe the reason that our model is better at summarization but not on QA tasks is due to the bias learned by BART, which we initialize our model from. In contrast to T5 which is pretrained to predict short spans and thus might be specialized at predicting factual targets like named entities, BART is extensively pretrained to decoding long and full sentences/paragraphs. 


\subsection{Performance Analysis on Input Lengths}
To further investigate the performance gains of our proposed model, we compare the performance of the proposed model against the base model as a function of source document length for two summarization datasets, namely SummScreen and TVMegaSite. 
To conduct our analysis, we divide the validation split of both the datasets into \emph{short} and \emph{long} documents.  
The cutoff length to separate the two groups is chosen such that approximately 75\% of the documents are classified as short documents. 
Figure~\ref{fig:len_comp} presents the results of this comparison.  
For both the datasets: (a) there's a performance drop for both the best and the base model for longer documents, and (b) the best model is better than the base model on all data splits. 
For SummScreen the performance gap between the best and the base model is bigger for \emph{long} documents than for \emph{short} documents -- relative ROUGE-L increase of 0.80\% and 3.96\% for \emph{short} and \emph{long} documents respectively. 
This suggests that the performance gains for the best model can be attributed to better long-context modeling.  
For TVMegaSite this trend of increasing performance gap between the best and the base model with an increase in document length still holds true, though the increase in performance gap is modest in comparison to the increase observed for SummScreen -- relative ROUGE-L increase of 2.43\% and 2.75\% for \emph{short} and \emph{long} documents respectively.

\section{Related Work}

\subsection{Efficient Long-Context Architectures} 

A long list of works propose to reduce the complexity of the attention layers of transformers. The simplest paradigm to achieve efficiency is to restrict each token's attending context to a subset of the whole sequences, e.g., Reformer~\citep{reformer} and the Routing transformer~\citep{routing} proposes hashing or clustering based attention, where each token only attends to tokens of a single bucket/cluster. Our model architecture is mostly influenced by previous work like Longformer~\citep{longformer}, BigBird~\citep{Bigbird} and ETC~\citep{ETC} that have demonstrated strong downstream performance. These models assume strong locality bias in language data and restrict each token's attending context to nearby tokens. In contrast to these works, we augment the block attention with pooling layers and study the effect of additional pretraining on long sequences. Other popular approach to tackle the efficiency bottleneck includes kernel-based~\citep{performer,rfa} and low-rank approximation~\citep{linformer} of the N$\times$N attention matrix. However, in contrast to local attention transformers, the effectiveness of these approximation approaches is yet to be validated in large models and downstream tasks. 

Apart from methods that solely modify transformer's attention calculations, several recent works propose alternative architectures that are free of the quadratic complexity over input length by design. For instance. Perceiver IO~\citep{perceiver_io} proposes to maintain a limited-length latent sequence instead of all tokens' hidden states, and only conduct attention computation on the latent space. In another notable work, \citet{s4} propose to use structured state-space models rooted in control theory to model the dependencies in long sequences. The model can be trained as a CNN with large kernel size, which is implemented efficiently with fast fourier transforms, and tested as a recurrent network that has $O(1)$ complexity. For these models to achieve reasonable performance on the downstream tasks we studied, it is necessary to pretrain them from scratch, which will require significant computation compared to our adapting approach. However, testing these emerging models on scaled settings and real-world long-sequence problems is indeed an important research problem and we hope to explore these alternative architectures in future work.

\subsection{Conditional Generation from Long Text}
To apply pretrained models to long-sequence tasks, early work~\citep{Bigbird, longformer} simply reuses parameters from models pretrained on short sequences and replaces the encoder full attention with sparse local attentions. While the models are not exposed to long sequences at pretraining time, they demonstrates consistent improvements over previous models that can only take truncated inputs. Complementary to local attentions, \citet{poolingformer} show that pooling layers can be inserted into a pretrained transformer at finetuning time and bring additional performance gains on summarization. Instead of relying on a single model that directly processes the whole input, \citet{dyle} proposes a two-stage extract-and-generate approach, where the extractor can leverage the supervision signal learned by the generator. However, despite the complicated training recipe, it does not bring consistent gains and underperforms our non-pretrain baselines. The most relevant work to ours is LongT5~\citep{longt5}, which adopts both global tokens as well as local attention, and pretrains the model with 4k text sequences from C4. Compared to LongT5, we augment local attentions with pooling layers and present a more comprehensive study on pretraining strategies. Without pretraining from scratch, we achieve stronger summarization performance. Concurrent to our work, ~\citet{phang2022investigating} also present an empirical study on adapting short-text models for long document summarization. While their study mostly focuses on the architecture aspect, we present additional analysis on the choices of pretraining corpus and learning objectives.

\section{Conclusion}
Through a comprehensive study on the effects of model architectures, training losses and pretraining dataset, we present an effective recipe to adapt existing pretrained text-to-text models for long-sequence NLP tasks. 
The resulting model sets new state-of-the-art on five long-sequence summarization tasks and achieves consistent gains on QA over local-attention models that simply reuse BART's parameters.
Apart from presenting a stronger checkpoint for finetuning on downstream tasks, we hope our findings in the study can provide insights for future works that aim to develop stronger long-sequence models for downstream tasks.

\section{Limitations}

Pretraining language models is a costly endeavor, and even more so in the case of long-context PLMs. 
Because of computational budget constraints, we only explored a limited space of the hyperparameter search space.  
\begin{itemize}
    \item We experiment with training on either just long document corpora or a pseudo long document corpora formed by concatenating short documents. 
    Future work can investigate using a combination of the two.   
    \item We have a surprising empirical finding that pretraining on \emph{pseudo} long documents formed by concatenating random documents of a short-document corpora (C4) outperforms both: (a) pretraining on actual long documents from a long-document corpora, and (b) pretraining on \emph{pseudo} long documents formed by concatenating related documents from the same short-document corpora. 
    Future work can investigate in more detail the reasons for these empirical gains, and also test these models on their discourse understanding.  
    \item Due to the human evaluation cost for long-context summarization tasks, we rely on automatic metrics which can be unreliable as suggested by prior work~\citep{kryscinski-etal-2019-neural, fabbri-etal-2021-summeval}.
\end{itemize}

\bibliography{anthology,custom}

\begin{thebibliography}{49}
\expandafter\ifx\csname natexlab\endcsname\relax\def\natexlab#1{#1}\fi

\bibitem[{Aghajanyan et~al.(2021)Aghajanyan, Shrivastava, Gupta, Goyal,
  Zettlemoyer, and Gupta}]{r3f}
Armen Aghajanyan, Akshat Shrivastava, Anchit Gupta, Naman Goyal, Luke
  Zettlemoyer, and Sonal Gupta. 2021.
\newblock \href {https://openreview.net/forum?id=OQ08SN70M1V} {Better
  fine-tuning by reducing representational collapse}.
\newblock In \emph{9th International Conference on Learning Representations,
  {ICLR} 2021, Virtual Event, Austria, May 3-7, 2021}. OpenReview.net.

\bibitem[{Ainslie et~al.(2020)Ainslie, Onta{\~{n}}{\'{o}}n, Alberti, Cvicek,
  Fisher, Pham, Ravula, Sanghai, Wang, and Yang}]{ETC}
Joshua Ainslie, Santiago Onta{\~{n}}{\'{o}}n, Chris Alberti, Vaclav Cvicek,
  Zachary Fisher, Philip Pham, Anirudh Ravula, Sumit Sanghai, Qifan Wang, and
  Li~Yang. 2020.
\newblock {ETC:} encoding long and structured inputs in transformers.
\newblock In \emph{{EMNLP} {(1)}}, pages 268--284. Association for
  Computational Linguistics.

\bibitem[{Beltagy et~al.(2020)Beltagy, Peters, and Cohan}]{longformer}
Iz~Beltagy, Matthew~E. Peters, and Arman Cohan. 2020.
\newblock \href {http://arxiv.org/abs/2004.05150} {Longformer: The
  long-document transformer}.
\newblock abs/2004.05150.

\bibitem[{Chen et~al.(2022)Chen, Chu, Wiseman, and Gimpel}]{summscreen}
Mingda Chen, Zewei Chu, Sam Wiseman, and Kevin Gimpel. 2022.
\newblock \href {https://doi.org/10.18653/v1/2022.acl-long.589}
  {{S}umm{S}creen: A dataset for abstractive screenplay summarization}.
\newblock In \emph{Proceedings of the 60th Annual Meeting of the Association
  for Computational Linguistics (Volume 1: Long Papers)}, pages 8602--8615,
  Dublin, Ireland. Association for Computational Linguistics.

\bibitem[{Choromanski et~al.(2021)Choromanski, Likhosherstov, Dohan, Song,
  Gane, Sarl{\'{o}}s, Hawkins, Davis, Mohiuddin, Kaiser, Belanger, Colwell, and
  Weller}]{performer}
Krzysztof Choromanski, Valerii Likhosherstov, David Dohan, Xingyou Song,
  Andreea Gane, Tam{\'{a}}s Sarl{\'{o}}s, Peter Hawkins, Jared~Quincy Davis,
  Afroz Mohiuddin, Lukasz Kaiser, David~Benjamin Belanger, Lucy~J. Colwell, and
  Adrian Weller. 2021.
\newblock Rethinking attention with performers.
\newblock In \emph{{ICLR}}. OpenReview.net.

\bibitem[{Clark et~al.(2020)Clark, Luong, Le, and Manning}]{electra}
Kevin Clark, Minh{-}Thang Luong, Quoc~V. Le, and Christopher~D. Manning. 2020.
\newblock {ELECTRA:} pre-training text encoders as discriminators rather than
  generators.
\newblock In \emph{{ICLR}}. OpenReview.net.

\bibitem[{Cohan et~al.(2018)Cohan, Dernoncourt, Kim, Bui, Kim, Chang, and
  Goharian}]{arxiv_pubmed}
Arman Cohan, Franck Dernoncourt, Doo~Soon Kim, Trung Bui, Seokhwan Kim, Walter
  Chang, and Nazli Goharian. 2018.
\newblock \href {https://doi.org/10.18653/v1/N18-2097} {A discourse-aware
  attention model for abstractive summarization of long documents}.
\newblock In \emph{Proceedings of the 2018 Conference of the North {A}merican
  Chapter of the Association for Computational Linguistics: Human Language
  Technologies, Volume 2 (Short Papers)}, pages 615--621, New Orleans,
  Louisiana. Association for Computational Linguistics.

\bibitem[{Dasigi et~al.(2021)Dasigi, Lo, Beltagy, Cohan, Smith, and
  Gardner}]{qasper}
Pradeep Dasigi, Kyle Lo, Iz~Beltagy, Arman Cohan, Noah~A. Smith, and Matt
  Gardner. 2021.
\newblock A dataset of information-seeking questions and answers anchored in
  research papers.
\newblock In \emph{NAACL}.

\bibitem[{Devlin et~al.(2019)Devlin, Chang, Lee, and
  Toutanova}]{devlin-etal-2019-bert}
Jacob Devlin, Ming-Wei Chang, Kenton Lee, and Kristina Toutanova. 2019.
\newblock \href {https://doi.org/10.18653/v1/N19-1423} {{BERT}: Pre-training of
  deep bidirectional transformers for language understanding}.
\newblock In \emph{Proceedings of the 2019 Conference of the North {A}merican
  Chapter of the Association for Computational Linguistics: Human Language
  Technologies, Volume 1 (Long and Short Papers)}, pages 4171--4186,
  Minneapolis, Minnesota. Association for Computational Linguistics.

\bibitem[{Fabbri et~al.(2021)Fabbri, Kry{\'s}ci{\'n}ski, McCann, Xiong, Socher,
  and Radev}]{fabbri-etal-2021-summeval}
Alexander~R. Fabbri, Wojciech Kry{\'s}ci{\'n}ski, Bryan McCann, Caiming Xiong,
  Richard Socher, and Dragomir Radev. 2021.
\newblock \href {https://doi.org/10.1162/tacl_a_00373} {{S}umm{E}val:
  Re-evaluating summarization evaluation}.
\newblock \emph{Transactions of the Association for Computational Linguistics},
  9:391--409.

\bibitem[{Gao et~al.(2021)Gao, Biderman, Black, Golding, Hoppe, Foster, Phang,
  He, Thite, Nabeshima, Presser, and Leahy}]{thepile}
Leo Gao, Stella Biderman, Sid Black, Laurence Golding, Travis Hoppe, Charles
  Foster, Jason Phang, Horace He, Anish Thite, Noa Nabeshima, Shawn Presser,
  and Connor Leahy. 2021.
\newblock The pile: An 800gb dataset of diverse text for language modeling.
\newblock \emph{arXiv}, abs/2101.00027.

\bibitem[{Gu et~al.(2021)Gu, Goel, and R{\'{e}}}]{s4}
Albert Gu, Karan Goel, and Christopher R{\'{e}}. 2021.
\newblock Efficiently modeling long sequences with structured state spaces.

\bibitem[{Guo et~al.(2022)Guo, Ainslie, Uthus, Ontanon, Ni, Sung, and
  Yang}]{longt5}
Mandy Guo, Joshua Ainslie, David Uthus, Santiago Ontanon, Jianmo Ni, Yun-Hsuan
  Sung, and Yinfei Yang. 2022.
\newblock \href {https://aclanthology.org/2022.findings-naacl.55} {{L}ong{T}5:
  {E}fficient text-to-text transformer for long sequences}.
\newblock In \emph{Findings of the Association for Computational Linguistics:
  NAACL 2022}, pages 724--736, Seattle, United States. Association for
  Computational Linguistics.

\bibitem[{Huang et~al.(2021)Huang, Cao, Parulian, Ji, and Wang}]{govreport}
Luyang Huang, Shuyang Cao, Nikolaus Parulian, Heng Ji, and Lu~Wang. 2021.
\newblock \href {https://doi.org/10.18653/v1/2021.naacl-main.112} {Efficient
  attentions for long document summarization}.
\newblock In \emph{Proceedings of the 2021 Conference of the North American
  Chapter of the Association for Computational Linguistics: Human Language
  Technologies}, pages 1419--1436, Online. Association for Computational
  Linguistics.

\bibitem[{Izacard et~al.(2021)Izacard, Caron, Hosseini, Riedel, Bojanowski,
  Joulin, and Grave}]{contriver}
Gautier Izacard, Mathilde Caron, Lucas Hosseini, Sebastian Riedel, Piotr
  Bojanowski, Armand Joulin, and Edouard Grave. 2021.
\newblock Towards unsupervised dense information retrieval with contrastive
  learning.
\newblock \emph{arXiv}, abs/2112.09118.

\bibitem[{Izacard and Grave(2021)}]{fid}
Gautier Izacard and Edouard Grave. 2021.
\newblock \href {https://doi.org/10.18653/v1/2021.eacl-main.74} {Leveraging
  passage retrieval with generative models for open domain question answering}.
\newblock In \emph{Proceedings of the 16th Conference of the European Chapter
  of the Association for Computational Linguistics: Main Volume}, pages
  874--880, Online. Association for Computational Linguistics.

\bibitem[{Jaegle et~al.(2021)Jaegle, Borgeaud, Alayrac, Doersch, Ionescu, Ding,
  Koppula, Zoran, Brock, Shelhamer et~al.}]{perceiver_io}
Andrew Jaegle, Sebastian Borgeaud, Jean-Baptiste Alayrac, Carl Doersch, Catalin
  Ionescu, David Ding, Skanda Koppula, Daniel Zoran, Andrew Brock, Evan
  Shelhamer, et~al. 2021.
\newblock Perceiver io: A general architecture for structured inputs \&
  outputs.
\newblock \emph{arXiv preprint arXiv:2107.14795}.

\bibitem[{Kitaev et~al.(2020)Kitaev, Kaiser, and Levskaya}]{reformer}
Nikita Kitaev, Lukasz Kaiser, and Anselm Levskaya. 2020.
\newblock Reformer: The efficient transformer.
\newblock In \emph{{ICLR}}. OpenReview.net.

\bibitem[{Ko{\v{c}}isk{\'y} et~al.(2018)Ko{\v{c}}isk{\'y}, Schwarz, Blunsom,
  Dyer, Hermann, Melis, and Grefenstette}]{narrativeqa}
Tom{\'a}{\v{s}} Ko{\v{c}}isk{\'y}, Jonathan Schwarz, Phil Blunsom, Chris Dyer,
  Karl~Moritz Hermann, G{\'a}bor Melis, and Edward Grefenstette. 2018.
\newblock \href {https://doi.org/10.1162/tacl_a_00023} {The {N}arrative{QA}
  reading comprehension challenge}.
\newblock \emph{Transactions of the Association for Computational Linguistics},
  6:317--328.

\bibitem[{Kryscinski et~al.(2019)Kryscinski, Keskar, McCann, Xiong, and
  Socher}]{kryscinski-etal-2019-neural}
Wojciech Kryscinski, Nitish~Shirish Keskar, Bryan McCann, Caiming Xiong, and
  Richard Socher. 2019.
\newblock \href {https://doi.org/10.18653/v1/D19-1051} {Neural text
  summarization: A critical evaluation}.
\newblock In \emph{Proceedings of the 2019 Conference on Empirical Methods in
  Natural Language Processing and the 9th International Joint Conference on
  Natural Language Processing (EMNLP-IJCNLP)}, pages 540--551, Hong Kong,
  China. Association for Computational Linguistics.

\bibitem[{Kry{\'s}ci{\'n}ski et~al.(2021)Kry{\'s}ci{\'n}ski, Rajani, Agarwal,
  Xiong, and Radev}]{booksum}
Wojciech Kry{\'s}ci{\'n}ski, Nazneen Rajani, Divyansh Agarwal, Caiming Xiong,
  and Dragomir Radev. 2021.
\newblock \href {http://arxiv.org/abs/2105.08209} {Booksum: A collection of
  datasets for long-form narrative summarization}.

\bibitem[{Lei(2021)}]{sru+}
Tao Lei. 2021.
\newblock \href {https://aclanthology.org/2021.emnlp-main.602} {When attention
  meets fast recurrence: Training language models with reduced compute}.
\newblock In \emph{Proceedings of the 2021 Conference on Empirical Methods in
  Natural Language Processing}, pages 7633--7648, Online and Punta Cana,
  Dominican Republic. Association for Computational Linguistics.

\bibitem[{Lewis et~al.(2020)Lewis, Liu, Goyal, Ghazvininejad, Mohamed, Levy,
  Stoyanov, and Zettlemoyer}]{lewis-etal-2020-bart}
Mike Lewis, Yinhan Liu, Naman Goyal, Marjan Ghazvininejad, Abdelrahman Mohamed,
  Omer Levy, Veselin Stoyanov, and Luke Zettlemoyer. 2020.
\newblock \href {https://doi.org/10.18653/v1/2020.acl-main.703} {{BART}:
  Denoising sequence-to-sequence pre-training for natural language generation,
  translation, and comprehension}.
\newblock In \emph{Proceedings of the 58th Annual Meeting of the Association
  for Computational Linguistics}, pages 7871--7880, Online. Association for
  Computational Linguistics.

\bibitem[{Liu et~al.(2022)Liu, Ni, Nan, Deb, Zhu, Awadallah, and
  Radev}]{pagesum}
Yixin Liu, Ansong Ni, Linyong Nan, Budhaditya Deb, Chenguang Zhu, Ahmed~H
  Awadallah, and Dragomir Radev. 2022.
\newblock Leveraging locality in abstractive text summarization.
\newblock \emph{arXiv preprint arXiv:2205.12476}.

\bibitem[{Mao et~al.(2022)Mao, Wu, Ni, Zhang, Zhang, Yu, Deb, Zhu, Awadallah,
  and Radev}]{dyle}
Ziming Mao, Chen~Henry Wu, Ansong Ni, Yusen Zhang, Rui Zhang, Tao Yu,
  Budhaditya Deb, Chenguang Zhu, Ahmed Awadallah, and Dragomir Radev. 2022.
\newblock \href {https://doi.org/10.18653/v1/2022.acl-long.118} {{DYLE}:
  Dynamic latent extraction for abstractive long-input summarization}.
\newblock In \emph{Proceedings of the 60th Annual Meeting of the Association
  for Computational Linguistics (Volume 1: Long Papers)}, pages 1687--1698,
  Dublin, Ireland. Association for Computational Linguistics.

\bibitem[{Pang et~al.(2022{\natexlab{a}})Pang, Nijkamp, Kry{\'s}ci{\'n}ski,
  Savarese, Zhou, and Xiong}]{topdown}
Bo~Pang, Erik Nijkamp, Wojciech Kry{\'s}ci{\'n}ski, Silvio Savarese, Yingbo
  Zhou, and Caiming Xiong. 2022{\natexlab{a}}.
\newblock Long document summarization with top-down and bottom-up inference.
\newblock \emph{arXiv preprint arXiv:2203.07586}.

\bibitem[{Pang et~al.(2022{\natexlab{b}})Pang, Parrish, Joshi, Nangia, Phang,
  Chen, Padmakumar, Ma, Thompson, He, and Bowman}]{quality}
Richard~Yuanzhe Pang, Alicia Parrish, Nitish Joshi, Nikita Nangia, Jason Phang,
  Angelica Chen, Vishakh Padmakumar, Johnny Ma, Jana Thompson, He~He, and
  Samuel Bowman. 2022{\natexlab{b}}.
\newblock \href {https://aclanthology.org/2022.naacl-main.391} {{Q}u{ALITY}:
  Question answering with long input texts, yes!}
\newblock In \emph{Proceedings of the 2022 Conference of the North American
  Chapter of the Association for Computational Linguistics: Human Language
  Technologies}, pages 5336--5358, Seattle, United States. Association for
  Computational Linguistics.

\bibitem[{Peng et~al.(2021)Peng, Pappas, Yogatama, Schwartz, Smith, and
  Kong}]{rfa}
Hao Peng, Nikolaos Pappas, Dani Yogatama, Roy Schwartz, Noah~A. Smith, and
  Lingpeng Kong. 2021.
\newblock Random feature attention.
\newblock In \emph{{ICLR}}. OpenReview.net.

\bibitem[{Phang et~al.(2022)Phang, Zhao, and Liu}]{phang2022investigating}
Jason Phang, Yao Zhao, and Peter~J Liu. 2022.
\newblock Investigating efficiently extending transformers for long input
  summarization.
\newblock \emph{arXiv preprint arXiv:2208.04347}.

\bibitem[{Raffel et~al.(2020)Raffel, Shazeer, Roberts, Lee, Narang, Matena,
  Zhou, Li, and Liu}]{t5}
Colin Raffel, Noam Shazeer, Adam Roberts, Katherine Lee, Sharan Narang, Michael
  Matena, Yanqi Zhou, Wei Li, and Peter~J. Liu. 2020.
\newblock Exploring the limits of transfer learning with a unified text-to-text
  transformer.
\newblock \emph{J. Mach. Learn. Res.}, 21:140:1--140:67.

\bibitem[{Rajpurkar et~al.(2016)Rajpurkar, Zhang, Lopyrev, and
  Liang}]{rajpurkar-etal-2016-squad}
Pranav Rajpurkar, Jian Zhang, Konstantin Lopyrev, and Percy Liang. 2016.
\newblock \href {https://doi.org/10.18653/v1/D16-1264} {{SQ}u{AD}: 100,000+
  questions for machine comprehension of text}.
\newblock In \emph{Proceedings of the 2016 Conference on Empirical Methods in
  Natural Language Processing}, pages 2383--2392, Austin, Texas. Association
  for Computational Linguistics.

\bibitem[{Roy et~al.(2021)Roy, Saffar, Vaswani, and Grangier}]{routing}
Aurko Roy, Mohammad Saffar, Ashish Vaswani, and David Grangier. 2021.
\newblock \href {https://doi.org/10.1162/tacl_a_00353} {Efficient content-based
  sparse attention with routing transformers}.
\newblock \emph{Transactions of the Association for Computational Linguistics},
  9:53--68.

\bibitem[{Shaham et~al.(2022)Shaham, Segal, Ivgi, Efrat, Yoran, Haviv, Gupta,
  Xiong, Geva, Berant, and Levy}]{scrolls}
Uri Shaham, Elad Segal, Maor Ivgi, Avia Efrat, Ori Yoran, Adi Haviv, Ankit
  Gupta, Wenhan Xiong, Mor Geva, Jonathan Berant, and Omer Levy. 2022.
\newblock \href {http://arxiv.org/abs/2201.03533} {Scrolls: Standardized
  comparison over long language sequences}.

\bibitem[{Tay et~al.(2022)Tay, Dehghani, Abnar, Chung, Fedus, Rao, Narang,
  Tran, Yogatama, and Metzler}]{scaling_laws}
Yi~Tay, Mostafa Dehghani, Samira Abnar, Hyung~Won Chung, William Fedus, Jinfeng
  Rao, Sharan Narang, Vinh~Q. Tran, Dani Yogatama, and Donald Metzler. 2022.
\newblock \href {https://doi.org/10.48550/arXiv.2207.10551} {Scaling laws vs
  model architectures: How does inductive bias influence scaling?}
\newblock \emph{CoRR}, abs/2207.10551.

\bibitem[{Tay et~al.(2020)Tay, Dehghani, Bahri, and Metzler}]{attention_survey}
Yi~Tay, Mostafa Dehghani, Dara Bahri, and Donald Metzler. 2020.
\newblock \href {http://arxiv.org/abs/2009.06732} {Efficient transformers: {A}
  survey}.
\newblock \emph{arXiv}, abs/2009.06732.

\bibitem[{Tiedemann(2016)}]{opensub}
J{\"o}rg Tiedemann. 2016.
\newblock \href {https://aclanthology.org/L16-1559} {Finding alternative
  translations in a large corpus of movie subtitle}.
\newblock In \emph{Proceedings of the Tenth International Conference on
  Language Resources and Evaluation ({LREC}'16)}, pages 3518--3522,
  Portoro{\v{z}}, Slovenia. European Language Resources Association (ELRA).

\bibitem[{Trinh and Le(2018)}]{stories}
Trieu~H. Trinh and Quoc~V. Le. 2018.
\newblock A simple method for commonsense reasoning.
\newblock \emph{arXiv}, abs/1806.02847.

\bibitem[{Vaswani et~al.(2017)Vaswani, Shazeer, Parmar, Uszkoreit, Jones,
  Gomez, Kaiser, and Polosukhin}]{Transformers}
Ashish Vaswani, Noam Shazeer, Niki Parmar, Jakob Uszkoreit, Llion Jones,
  Aidan~N Gomez, \L~ukasz Kaiser, and Illia Polosukhin. 2017.
\newblock \href
  {https://proceedings.neurips.cc/paper/2017/file/3f5ee243547dee91fbd053c1c4a845aa-Paper.pdf}
  {Attention is all you need}.
\newblock In \emph{Advances in Neural Information Processing Systems},
  volume~30. Curran Associates, Inc.

\bibitem[{Vig et~al.(2022)Vig, Fabbri, Kryscinski, Wu, and Liu}]{qmsum_trick}
Jesse Vig, Alexander Fabbri, Wojciech Kryscinski, Chien-Sheng Wu, and Wenhao
  Liu. 2022.
\newblock \href {https://aclanthology.org/2022.findings-naacl.109} {Exploring
  neural models for query-focused summarization}.
\newblock In \emph{Findings of the Association for Computational Linguistics:
  NAACL 2022}, pages 1455--1468, Seattle, United States. Association for
  Computational Linguistics.

\bibitem[{Wang et~al.(2019)Wang, Singh, Michael, Hill, Levy, and Bowman}]{glue}
Alex Wang, Amanpreet Singh, Julian Michael, Felix Hill, Omer Levy, and
  Samuel~R. Bowman. 2019.
\newblock {GLUE:} {A} multi-task benchmark and analysis platform for natural
  language understanding.
\newblock In \emph{{ICLR} (Poster)}. OpenReview.net.

\bibitem[{Wang et~al.(2020)Wang, Li, Khabsa, Fang, and Ma}]{linformer}
Sinong Wang, Belinda~Z. Li, Madian Khabsa, Han Fang, and Hao Ma. 2020.
\newblock Linformer: Self-attention with linear complexity.
\newblock \emph{arXiv}, abs/2006.04768.

\bibitem[{Xiong et~al.(2022)Xiong, Oguz, Gupta, Chen, Liskovich, Levy, Yih, and
  Mehdad}]{SimpleLA}
Wenhan Xiong, Barlas Oguz, Anchit Gupta, Xilun Chen, Diana Liskovich, Omer
  Levy, Scott Yih, and Yashar Mehdad. 2022.
\newblock \href {https://aclanthology.org/2022.naacl-main.144} {Simple local
  attentions remain competitive for long-context tasks}.
\newblock In \emph{Proceedings of the 2022 Conference of the North American
  Chapter of the Association for Computational Linguistics: Human Language
  Technologies}, pages 1975--1986, Seattle, United States. Association for
  Computational Linguistics.

\bibitem[{Zaheer et~al.(2020)Zaheer, Guruganesh, Dubey, Ainslie, Alberti,
  Onta{\~{n}}{\'{o}}n, Pham, Ravula, Wang, Yang, and Ahmed}]{Bigbird}
Manzil Zaheer, Guru Guruganesh, Kumar~Avinava Dubey, Joshua Ainslie, Chris
  Alberti, Santiago Onta{\~{n}}{\'{o}}n, Philip Pham, Anirudh Ravula, Qifan
  Wang, Li~Yang, and Amr Ahmed. 2020.
\newblock Big bird: Transformers for longer sequences.
\newblock In \emph{NeurIPS}.

\bibitem[{Zellers et~al.(2019)Zellers, Holtzman, Rashkin, Bisk, Farhadi,
  Roesner, and Choi}]{Realnews}
Rowan Zellers, Ari Holtzman, Hannah Rashkin, Yonatan Bisk, Ali Farhadi,
  Franziska Roesner, and Yejin Choi. 2019.
\newblock Defending against neural fake news.
\newblock In \emph{NeurIPS}, pages 9051--9062.

\bibitem[{Zhang et~al.(2021)Zhang, Gong, Shen, Li, Lv, Duan, and
  Chen}]{poolingformer}
Hang Zhang, Yeyun Gong, Yelong Shen, Weisheng Li, Jiancheng Lv, Nan Duan, and
  Weizhu Chen. 2021.
\newblock Poolingformer: Long document modeling with pooling attention.
\newblock In \emph{{ICML}}, volume 139 of \emph{Proceedings of Machine Learning
  Research}, pages 12437--12446. {PMLR}.

\bibitem[{Zhang et~al.(2020)Zhang, Zhao, Saleh, and Liu}]{pegasus}
Jingqing Zhang, Yao Zhao, Mohammad Saleh, and Peter~J. Liu. 2020.
\newblock Pegasus: Pre-training with extracted gap-sentences for abstractive
  summarization.
\newblock In \emph{Proceedings of the 37th International Conference on Machine
  Learning}, ICML'20. JMLR.org.

\bibitem[{Zhong et~al.(2022)Zhong, Liu, Xu, Zhu, and Zeng}]{dialogLM}
Ming Zhong, Yang Liu, Yichong Xu, Chenguang Zhu, and Michael Zeng. 2022.
\newblock Dialoglm: Pre-trained model for long dialogue understanding and
  summarization.
\newblock \emph{AAAI}.

\bibitem[{Zhong et~al.(2021)Zhong, Yin, Yu, Zaidi, Mutuma, Jha, Awadallah,
  Celikyilmaz, Liu, Qiu, and Radev}]{qmsum}
Ming Zhong, Da~Yin, Tao Yu, Ahmad Zaidi, Mutethia Mutuma, Rahul Jha,
  Ahmed~Hassan Awadallah, Asli Celikyilmaz, Yang Liu, Xipeng Qiu, and Dragomir
  Radev. 2021.
\newblock \href {https://doi.org/10.18653/v1/2021.naacl-main.472} {{QMS}um: A
  new benchmark for query-based multi-domain meeting summarization}.
\newblock In \emph{Proceedings of the 2021 Conference of the North American
  Chapter of the Association for Computational Linguistics: Human Language
  Technologies}, pages 5905--5921, Online. Association for Computational
  Linguistics.

\bibitem[{Zhu et~al.(2021)Zhu, Liu, Mei, and Zeng}]{mediasum}
Chenguang Zhu, Yang Liu, Jie Mei, and Michael Zeng. 2021.
\newblock \href {https://doi.org/10.18653/v1/2021.naacl-main.474}
  {{M}edia{S}um: A large-scale media interview dataset for dialogue
  summarization}.
\newblock In \emph{Proceedings of the 2021 Conference of the North American
  Chapter of the Association for Computational Linguistics: Human Language
  Technologies}, pages 5927--5934, Online. Association for Computational
  Linguistics.

\end{thebibliography}
\bibliographystyle{acl_natbib}

\appendix

\newpage
\section{Appendix}
\label{sec:appendix}

\paragraph{Build the linked C4 corpus} We attempt to use text retrieval techniques to assemble long text sequences with the hope that the model can learn more long-range dependencies from linked relevant documents. We first encode all the documents into dense vectors with the Contriver~\citep{contriver} encoder. For documents that have more than 512 tokens, we use primary sentences~\citep{pegasus} as the input to the encoder. Directly retrieving documents from the whole index (340M vectors) is prohibitive in terms of computation cost. We follow the idea inverted indices, we first k-means to get 256 clusters of documents and then assemble long sequences within each cluster. Starting from each documents, we concatenate it with its top-k nearest neighbors until the length exceeds certain threshold. To avoid repeated documents, we enforce that each documents can appear in at most 2 sequences.

\begin{figure}[t]
\centering
\includegraphics[width=0.8\linewidth]{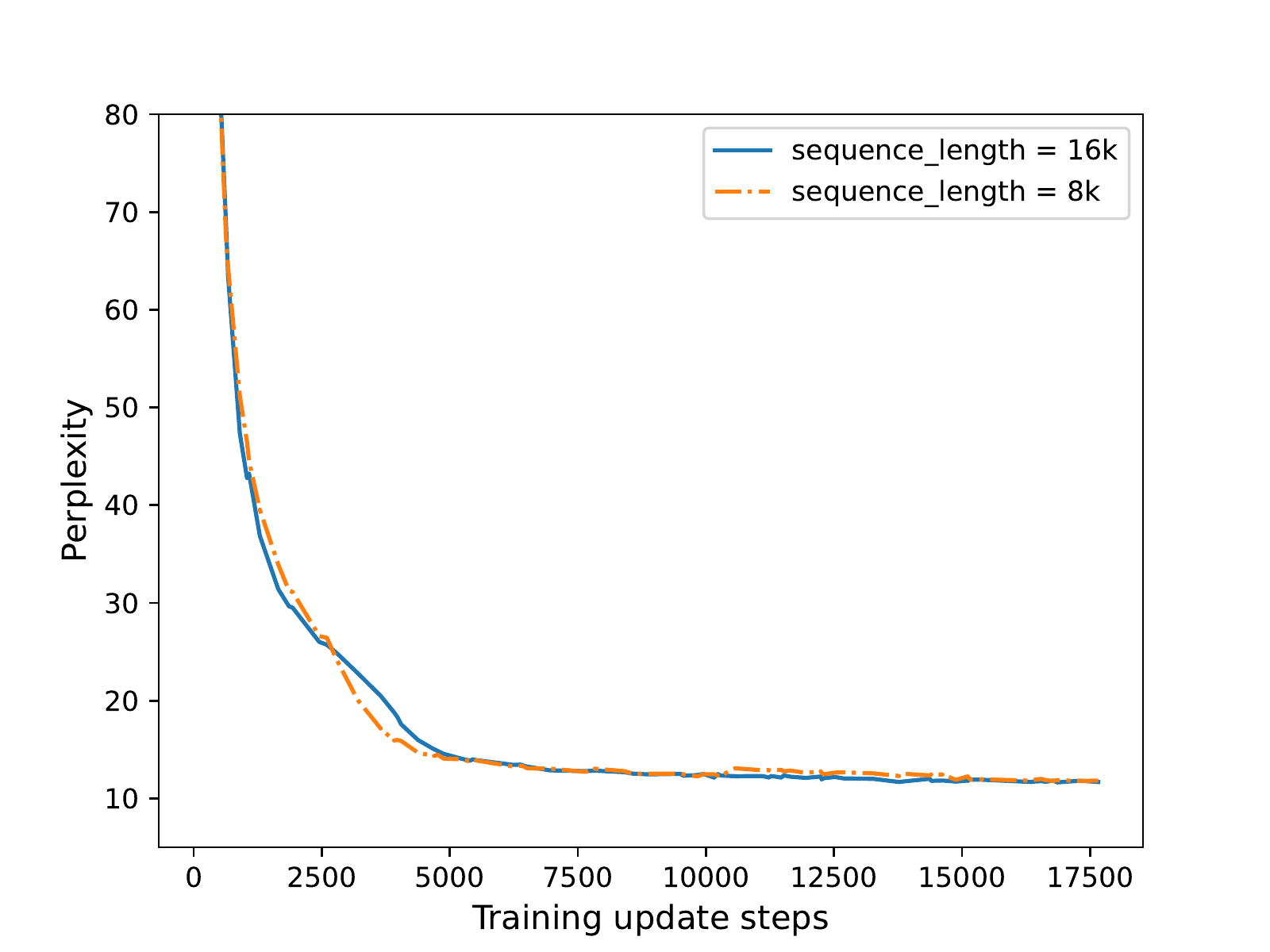}
\caption{Training curves with 8k/16k sequence lengths. Pretraining with different sequence lengths shows similar level of data efficiency.} 
\label{fig:pretrain_lens}
\end{figure}

\paragraph{Hyperparameters}
We use a fixed set of hyperparameters for pretraining: we set the learning rate to be $1e-4$, the weight decay coefficient to be $0.01$ and applies polynomial decay with $500$ warm up steps; we use a batch size of $256$ (16,384 tokens per sample) and fix the random seed to $42$. The hyperparameter grids for the downstream tasks are shown in Table~\ref{tab:finetune_hyper}.

\begin{table*}[h]
\small
\centering
    \begin{tabular}{l|l|c|c|l|c}
    \toprule
    Downstream Task & learning rate & batch size & max epoch & dropout & warmup steps (polynomial lr decay) \\
    \midrule
    arXiv &  1e-4, 3e-4, 4e-4 & 128 & 8 & 0, 0.1 & 200 \\
    \midrule
    GovReport &  5e-5, 3e-4, 4e-4 & 128 & 70 & 0, 0.1 & 200\\
    \midrule
    PubMed, BookSum &  3e-4, 4e-4 & 64 & 60 & 0, 0.1 & 200 \\
    \midrule 
    SummScreen & 5e-5, 3e-5, 1e-4 & 64 & 130 & 0, 0.1 & 200, 500, 1000 \\
    \midrule
    Qasper, QMSum, Quality & 1e-4, 5e-5, 3e-5 & 32, 64 & 150 & 0, 0.1 & 100, 200\\
    \midrule 
    NarrativeQA & 5e-5, 3e-5 & 64 & 8 & 0, 0.1 & 200\\
     \bottomrule
    \end{tabular}
\caption{Hyperparamter grid for downstream task finetuning. We use Adam optimizer ($\beta$ = (0.9, 0.999), $\epsilon$ = 1e-6)  for all tasks.}
\label{tab:finetune_hyper}
\end{table*}

\begin{table*}[h]
\small
\centering
    \begin{tabular}{l|l}
    \toprule
    Downstream Task & generation parameters \\
    \midrule
    arXiv &  beam: 4, max\_len: 300, min\_len: 50, length\_penalty: 5.0, no\_repeat\_ngram: 3\\
    \midrule
    GovReport & beam: 4, max\_len: 740, min\_len: 50, length\_penalty: 4.0, no\_repeat\_ngram: 3\\
    \midrule 
    PubMed &  beam: 4, max\_len: 400, min\_len: 40, length\_penalty: 4.0, no\_repeat\_ngram: 3\\
    \midrule
    BookSum & beam: 4, max\_len: 550, min\_len: 20, length\_penalty: 4.0, no\_repeat\_ngram: 3\\
    \midrule 
    SummScreen-FD  & beam: 4, max\_len: 300, min\_len: 50, length\_penalty: 4.0, no\_repeat\_ngram: 3\\
    \midrule
    SummScreen-TVM & beam: 4, max\_len: 640, min\_len: 50, length\_penalty: 5.0, no\_repeat\_ngram: 3\\
    \midrule
    Qasper  & beam: 4, max\_len: 80, length\_penalty: 1.0, no\_repeat\_ngram: 3\\
    \midrule 
    NarrativeQA & beam: 4, max\_len: 20, length\_penalty: 3.0, no\_repeat\_ngram: 3 \\
    \midrule
    QMSum & beam: 4, max\_len: 256, min\_len: 40, length\_penalty: 4.0, no\_repeat\_ngram: 3\\
    \midrule
    QuALITY & beam: 4, max\_len: 50, length\_penalty: 3.0, no\_repeat\_ngram: 3 \\
     \bottomrule
    \end{tabular}
\caption{Generation parameters for each task.}
\label{tab:generation_hyper}
\end{table*}

\end{document}